\documentclass[conference,numbers]{article}

\PassOptionsToPackage{numbers, compress}{natbib}

\usepackage[dblblindworkshop, final]{neurips_2025}

\usepackage[utf8]{inputenc} 
\usepackage[T1]{fontenc}    
\usepackage{hyperref}       
\usepackage{url}            
\usepackage{booktabs}       
\usepackage{amsfonts}       
\usepackage{nicefrac}       
\usepackage{microtype}      
\usepackage{xcolor}         

\usepackage{graphicx}
\usepackage{amsmath}
\usepackage{tabularx}
\usepackage{multirow}
\usepackage{comment}

 \usepackage{tcolorbox}
\usepackage{listings}
\usepackage{longtable}
\usepackage{booktabs}
\usepackage{colortbl} 
\usepackage{soul}
\usepackage{amsmath,amssymb,amsfonts} 
\usepackage{dsfont}                   
\usepackage{algorithm}                
\usepackage[noend]{algpseudocode}

\usepackage{graphicx,array,booktabs}
\newcolumntype{C}[1]{>{\centering\arraybackslash}m{#1}} 
\newcommand{\rothead}[1]{\rotatebox[origin=c]{90}{\parbox{1.7cm}{\centering #1}}}

\newcommand{\rotheadW}[2][2.8cm]{
  \rotatebox[origin=c]{90}{\parbox{#1}{\centering #2}}}

\tcbset{
  promptbox/.style={
    colback=black!0,  
    coltext=black,    
    colframe=black,   
    arc=0mm,
    boxrule=1pt,      
    left=1mm,
    right=1mm,
    top=1mm,
    bottom=1mm,
  }
}

\title{
Spatial Reasoning in Foundation Models: Benchmarking Object-Centric Spatial Understanding
}
\workshoptitle{SPACE in Vision, Language, and Embodied AI}

\author{
    Vahid Mirjalili\thanks{Equal contributions.}~\thanks{All authors are from Walmart Global Tech, USA}
    \And
    Ramin Giahi\footnotemark[1]
    \And
    Sriram Kollipara\footnotemark[1]
    \And
    Akshay Kekuda\footnotemark[1]
    \And
    Kehui Yao\footnotemark[1]
    \And
    Kai Zhao\footnotemark[1]~\thanks{Corresponding author: \texttt{kai.zhao@walmart.com}} 
    \And
    Jianpeng Xu
    \And
    Kaushiki Nag
    \And
    Sinduja Subramaniam
    \And
    Topojoy Biswas
    \And
    Evren Korpeoglu
    \And
    Kannan Achan
}

\begin{document}

\maketitle

\begin{abstract}
Spatial understanding is a critical capability for vision foundation models. While recent advances in large vision models or vision–language models (VLMs) have expanded recognition capabilities, most benchmarks emphasize localization accuracy rather than whether models capture how objects are arranged and related within a scene. This gap is consequential: effective scene understanding requires not only identifying objects, but reasoning about their relative positions, groupings, and depth. In this paper, we present a systematic benchmark for object-centric spatial reasoning in foundation models. Using a controlled synthetic dataset, we evaluate state-of-the-art vision models (e.g., GroundingDINO, Florence-2, OWLv2) and large VLMs (e.g., InternVL, LLaVA, GPT-4o) across three tasks: spatial localization, spatial reasoning, and downstream retrieval tasks. We find a stable trade-off: detectors such as GroundingDINO and OWLv2 deliver precise boxes with limited relational reasoning, while VLMs like SmolVLM and GPT-4o provide coarse layout cues and fluent captions but struggle with fine-grained spatial context. Our study highlights the gap between localization and true spatial understanding, and pointing toward the need for spatially-aware foundation models in the community. 
\end{abstract}

\section{Introduction}
Understanding and reasoning about spatial relationships between objects is a core challenge for vision–language systems and embodied AI \cite{mindgap2025}. Despite progress in object detection~\cite{yolov12016,fasterrcnn2016} and open-vocabulary recognition~\cite{vild2021,owlvit2022}, most benchmarks emphasize localization, identifying, and bounding objects~\cite{2025howwell,geobench2024,mindgap2025}, rather than contextual spatial reasoning. Yet, applications from embodied interaction to e-commerce recommendation~\cite{pinnerformer2022,mmrec2024} require models to detect objects and interpret how they are arranged and functionally related within a scene. In shopping images, for example, a sofa may appear with a coffee table, rug, and lamp. A useful system must capture relative position, containment, and grouping: a sofa next to a coffee table implies different recommendations than a sofa isolated against a wall. Without such context, results risk incoherence and poor alignment with user intent.

State-of-the-art vision models (e.g., GroundingDINO~\cite{gd1pt52024}, Florence-2~\cite{florence22024}, OWLv2~\cite{owlv22023}) and large VLMs (e.g., InternVL~\cite{internvl2024}, LLaVA~\cite{llava2023}, GPT-4o~\cite{gpt4o2024}) broaden what can be localized in complex scenes, but their spatial reasoning ability remains underexplored. A model that detects both a ``sofa'' and a ``lamp'' may still fail to infer their relative depth, errors that harm multi-item recommendation, scene retrieval, and embodied planning.
We present a systematic benchmarking study of spatial reasoning in vision models and VLMs on a synthetic shopping-scene data. Our contribution lies in the evaluation protocol and analysis rather than a new dataset: we probe whether state-of-the-art systems can (i) localize a focal product in clutter, (ii) capture its spatial relations with surrounding items, and (iii) leverage these relations for retrieval and recommendation via a unified detection-to-retrieval pipeline and spatial-localization.

Our results show a consistent gap: task-specific (e.g., object detection) vision models such as GroundingDINO~\cite{gd1pt52024}  achieve high localization precision, but lack spatial reasoning capability. 
On the other hand, large VLMs like  GPT-4o~\cite{gpt4o2024} produce descriptive captions and coarse layouts yet underperform when fine-grained spatial context is required. These results reveal a persistent divide between precise localization and true spatial understanding. Our benchmark surfaces this gap and provides standardized tasks and metrics, establishing a foundation for developing spatially aware foundation models that unify detection accuracy with contextual understanding. We release complete details of the benchmark, covering data generation, evaluation protocol, prompts, metrics, and failure analyses, to ensure reproducibility and extension to other domains.

\section{Experiments}

\subsection{Datasets}

To benchmark spatial reasoning capability of vision models and VLMs, we build a unique synthetic dataset where both localization and relevance are known by construction, allowing us to control viewpoint, clutter, and scene context.
Our synthetic dataset spans across nine furniture categories: \emph{Bed}, \emph{Chair}, \emph{Cabinet}, \emph{Desk}, \emph{Dresser}, \emph{Planter}, \emph{Shelf}, \emph{Sofa}, and \emph{Vase}. Images are created by rendering 3D products with random rotation, shift, and scale, to increase variety. The rendered images are then composited with background scenes. For each product, we split renders into database (DB) images that contain frontal views and query images that contain more angled views. Details of the data-generation pipeline are given in Appendix~\ref{appendix:data-gen}.
The synthesized dataset has roughly $11\mathrm{k}$ images, split into $3\mathrm{k}$ DB images and remaining $8\mathrm{k}$ are served as query images. For retrieval, each query is paired with its matching DB image from the same product, which we use as ground-truth relevance. The statistics per category are reported in the Appendix Table~\ref{tab:dataset-summary}.

\subsection{Benchmark Models}
We evaluate 14 models grouped into two families: 
\begin{itemize}
    \item Task-specific (e.g., object detection) vision models include D-FINE\cite{dfine2024}, OWL-ViT\cite{owlvit2022}, OWLv2\cite{owlv22023}, Florence-2\cite{florence2021}, GroundingDINO\cite{gd1pt52024}, and (LISA-7B/13B)\cite{lisa2024}. 
    \item General-purpose VLMs which include SmolVLM\cite{smolvlm2025}, InternVL\cite{internvl2024}, LLaVA\cite{llava1pt52024}, LLaVA-OneVision\cite{llavaonevision2024}, LLaVA-Next\cite{llavanext2024}, Gemini~2.5\cite{gemini2pt5pro2025}, and GPT-4o\cite{gpt4o2024}.
\end{itemize}
We run three experiment sets: (i) spatial localization, (ii) spatial reasoning, and (iii) image-based retrieval. For spatial localization, we extract the focal product's bounding box using both vision models and VLMs. Spatial reasoning includes two tasks: predicting the focal object's position on a coarse grid (with $2{\times}2$ and $3{\times}3$ settings) and predicting coarse depth as front vs.\ back (VLM-only). For retrieval, we compute VL-CLIP~\cite{vlclip2025} embeddings over crops from each method's predicted boxes on query images, and search an HNSW~\cite{hnsw2018} index built on DB-image embeddings. 
Full setup and metrics are given in Appendix~\ref{appendix:eval-protocol}.

\subsection{Main Results}

In general, for spatial localization, reasoning, and retrieval experiments, task-specific vision models outperform general-purpose VLMs by a clear margin. GroundingDINO is the strongest overall, leading both spatial localization and image-based retrieval. Among VLMs, InternVL and LLaVA variants are the most competitive, yet they still underperform the best vision models consistently across spatial localization, spatial reasoning, and retrieval.

Table~\ref{tab:grid-combined} reports spatial localization performance for coarse grid-cell prediction using accuracy, macro-$\mathrm{F}_1$, and $\mathrm{MCC}$ metrics. Among task-specific vision models, GroundingDINO is consistently strongest, while LISA-7B ranks lowest in this group. Among the VLMs, LLaVA-OneVision and InternVL are the top performers, whereas GPT-4o and Gemini~2.5 show the lowest performance overall. Comparing across groups, vision models  clearly outperform VLMs on both grid settings, and the performance gap widens as the grid becomes smaller.

\begin{figure*}[h!]
    \centering
    \includegraphics[width=0.85\linewidth]{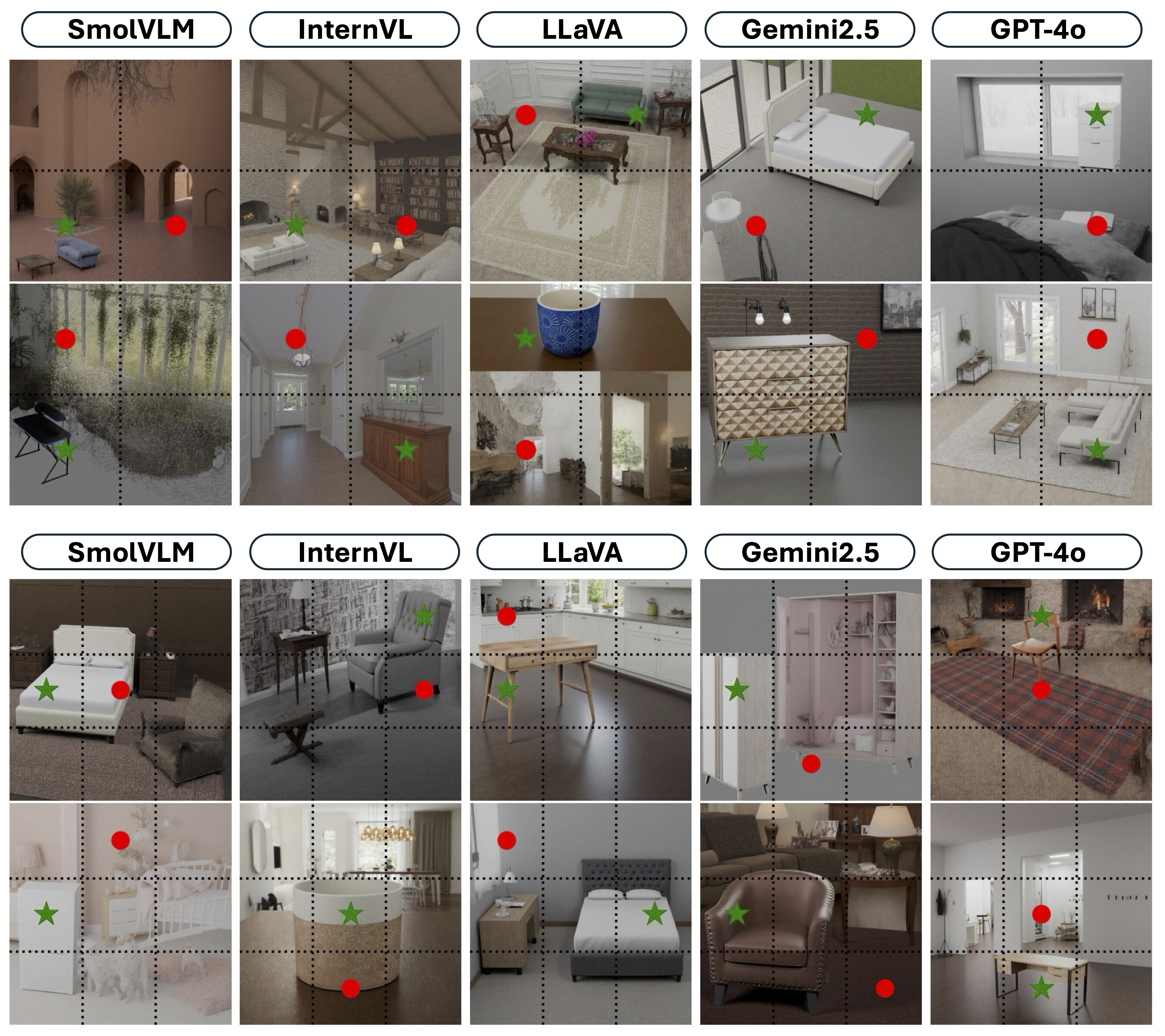}
    \caption{\textbf{Spatial localization failures of VLMs.}
    The GT cell is indicated by a \textcolor{black}{\emph{green star}}, and the predicted cell by a \textcolor{black}{\emph{red circle}} centered in the chosen cell.
    }
    \label{fig:grid-failures}
\end{figure*}

Figure~\ref{fig:grid-failures} illustrates typical spatial localization errors for VLMs. 
Across task-specific vision models, failures are dominated by box-placement issues (shifted or mis-sized extents) in cluttered or low-contrast scenes; while category confusions are comparatively rare. 
On the other hand, failures in VLMs are more often due to coarse spatial grounding, yielding off-center or overly loose boxes. This observation supports weaker box-level supervision in VLMs relative to vision models.

\begin{table}[t]
\caption{Spatial localization results for vision models and VLMs, predicting the location of the focal object in an image on a coarse grid}
\label{tab:grid-combined}
\centering
\scriptsize
\setlength{\tabcolsep}{4pt}
\resizebox{\textwidth}{!}{
\begin{tabular}{l l c c c c c c}
\toprule
& & \multicolumn{3}{c}{2$\times$2 grid} & \multicolumn{3}{c}{3$\times$3 grid} \\
\cmidrule(lr){3-5} \cmidrule(lr){6-8}
 & Model & Acc~$\uparrow$ & ${F_1}^{\mathrm{macro}}$~$\uparrow$ & MCC~$\uparrow$
         & Acc~$\uparrow$ & ${F_1}^{\mathrm{macro}}$~$\uparrow$ & MCC~$\uparrow$ \\
\midrule
\multirow{7}{*}{\shortstack[l]{\rothead{Task-specific\\vision models}}}
& GroundingDINO-1.5 & \textbf{0.816} & \textbf{0.815} & \textbf{0.754} & \textbf{0.799} & \textbf{0.759} & \textbf{0.742} \\
& Florence2-base    & \underline{0.808} & \underline{0.808} & \underline{0.745} & \underline{0.797} & \underline{0.754} & \underline{0.740} \\
& D-FINE            & 0.799 & 0.799 & 0.732 & 0.789 & 0.742 & 0.726 \\
& OWLv2-base-16     & 0.793 & 0.792 & 0.724 & 0.764 & 0.715 & 0.702 \\
& LISA-13B-Llama    & 0.785 & 0.785 & 0.713 & 0.750 & 0.690 & 0.684 \\
& OWL-ViT-base-32   & 0.759 & 0.759 & 0.679 & 0.728 & 0.677 & 0.655 \\
& LISA-7B           & 0.733 & 0.732 & 0.645 & 0.694 & 0.628 & 0.614 \\
\midrule
\multirow{7}{*}{\shortstack{\rothead{General-purpose\\VLMs}}}
& InternVL3-8B        & \underline{0.643} & \underline{0.640} & \underline{0.530} & \textbf{0.565} & \textbf{0.494} & \textbf{0.452} \\
& LLaVA-OneVision     & \textbf{0.645} & \textbf{0.643} & \textbf{0.538} & \underline{0.548} & \underline{0.390} & \underline{0.400} \\
& LLaVA-Next          & 0.551 & 0.544 & 0.415 & 0.429 & 0.239 & 0.244 \\
& LLaVA-1.5-7B        & 0.492 & 0.424 & 0.349 & 0.055 & 0.026 & 0.024 \\
& SmolVLM2-2.2B-Inst. & 0.422 & 0.360 & 0.253 & 0.355 & 0.106 & 0.071 \\
& Gemini~2.5-Pro      & 0.262 & 0.262 & 0.016 & 0.241 & 0.120 & 0.017 \\
& GPT-4o              & 0.252 & 0.199 & 0.003 & 0.315 & 0.088 & 0.015 \\
\bottomrule
\end{tabular}}
\end{table}

\begin{table*}[h!]
\caption{Assessing spatial reasoning capability of VLMs on predicting coarse-depth ordering of the focal object in an image.}
\label{tab:frontback-results}
\centering
\small
\setlength{\tabcolsep}{7pt}
\resizebox{0.8\textwidth}{!}
{%
\begin{tabular}{lcccc}
\toprule
Model & Acc~$\uparrow$ & Prec~$\uparrow$ & Rec~$\uparrow$ & F1~$\uparrow$ \\
\midrule
InternVL3-8B                 & \textbf{0.874} & 0.895 & 0.949 & \textbf{0.921} \\
SmolVLM2-2.2B-Instruct       & \underline{0.854} & 0.863 & \underline{0.964} & \underline{0.911} \\
LLaVA-OneVision              & 0.832 & 0.823 & \textbf{0.998} & 0.902 \\
LLaVA-1.5-7B                 & 0.745 & 0.907 & 0.748 & 0.820 \\
LLaVA-Next                   & 0.634 & \textbf{0.996} & 0.530 & 0.692 \\
Gemini~2.5-Pro               & 0.677 & 0.870 & 0.697 & 0.774 \\
GPT-4o                       & 0.455 & \underline{0.931} & 0.322 & 0.478 \\
\bottomrule
\end{tabular}}
\end{table*}

\subsection{Spatial Reasoning Capability of VLMs}

\begin{figure*}[h!]
\centering
\includegraphics[width=0.78\textwidth]{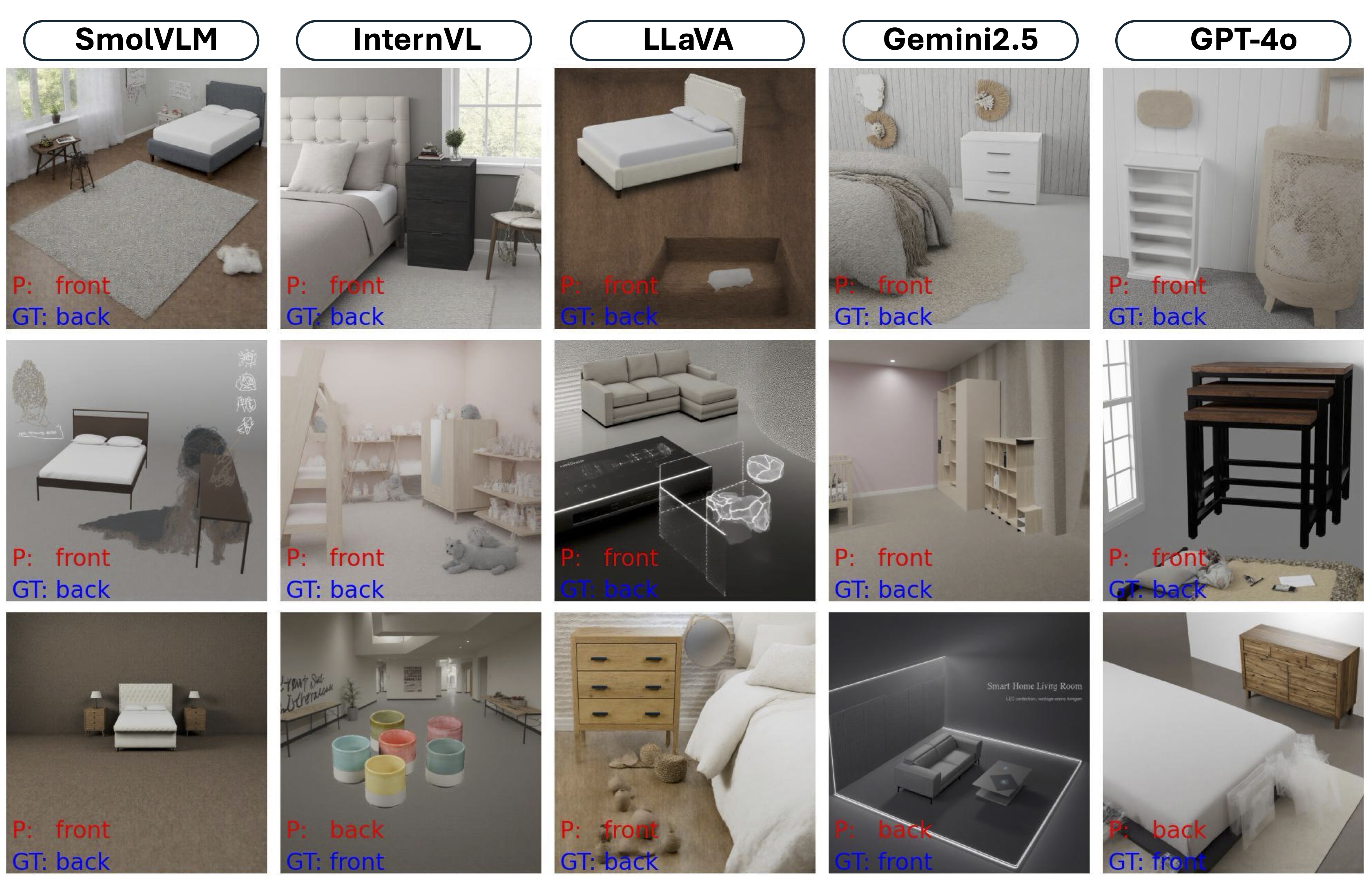}
\caption{Spatial reasoning failures of VLMs. Annotations: P is what the VLM thinks about the relative position of the item, GT is the ground-truth coarse depth of the item.}
\label{fig:frontback-failures}
\end{figure*}

We further analyze the spatial reasoning capability of VLMs for context-aware scene understanding, by evaluating coarse depth ordering of the focal object relative to its surrounding environment. For this experiment, VLMs are prompted for \emph{front-vs-back} classification (constrained to respond with \texttt{front} or \texttt{back}). This experiment assesses whether the model can determine if the focal object lies in the foreground (front) or background (back) of the overall scene by considering the relative depth and layering of objects. We report accuracy, precision, recall, and $\mathrm{F}_1$ in Table~\ref{tab:frontback-results}. 
The results show that InternVL is the best-performing model for coarse-depth prediction, achieving the highest accuracy and $\mathrm{F}_1$ score. While SmolVLM also shows strong performance, other models exhibit significant trade-offs between precision and recall. LLaVA-OneVision achieves near-perfect recall, indicating it correctly identifies almost every background object, though its precision is lower. In contrast, LLaVA-Next has extremely high precision but low recall. Performance drops notably for the last two models, with GPT-4o showing the weakest results, particularly in recall  and $\mathrm{F}_1$ score. For visualization purpose, we show failed cases in Figure \ref{fig:frontback-failures}. 

\clearpage

{\small
\bibliographystyle{ieee_fullname}
\bibliography{ref}
}

\clearpage
\appendix

\section*{\centering \normalfont\Large\bfseries Spatial Reasoning in Foundation Models: \\Benchmarking Object-Centric Spatial Understanding\\APPENDIX}

\vspace{1em} 
\addcontentsline{toc}{section}{Benchmarking Spatial Reasoning in Object-Centric Detection: A Study on Context-Aware Scene Understanding}

\section{Data Generation}
\label{appendix:data-gen}
\subsection{Data Generation Piepline}
\label{appendix:data-gen-pipeline}

Existing public datasets rarely pair precise box-level ground truth with product-level relevance for the same scenes. To address this, we construct a synthetic dataset in which both localization and relevance are known by construction, allowing controlled variation in viewpoint, clutter, and context.

\begin{figure}[h!]
    \centering
    \includegraphics[width=\linewidth]{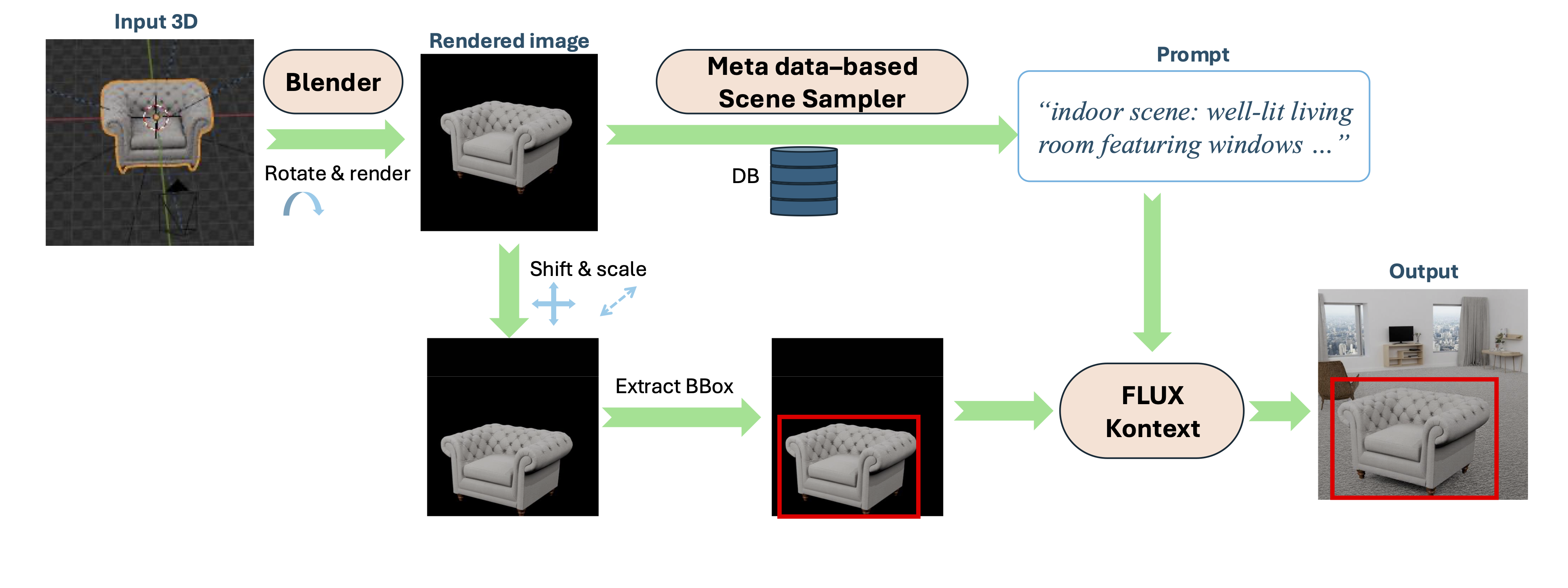}
    \caption{Data generation pipeline: generating synthetic product-in-scene images from a product's 3D asset based on a textual scene description sampled from a database.}
    \label{fig:data-gen-pipeline}
\end{figure}

We designed a data-generation pipeline to synthesize product–in–scene images by compositing 3D product renders into text-conditioned backgrounds (see Fig.~\ref{fig:data-gen-pipeline}). For each environment type (e.g., \emph{living room}, \emph{bedroom}, \emph{patio}), we generate short textual scene descriptions using GPT-4 and split them so that database (DB) and query images use disjoint descriptions. Each product’s 3D asset is rendered in Blender~\cite{blender2013} under random rotations to obtain a clean foreground mask; we apply random scale/shift augmentations and export the exact 2D GT bounding box from the transformed mask. The augmented foreground and a sampled scene description are fed to FLUX-Kontext~\cite{flux1kontext2025} to generate photorealistic composites. 
 
\begin{figure*}
    \centering
    \includegraphics[width=0.98\linewidth]{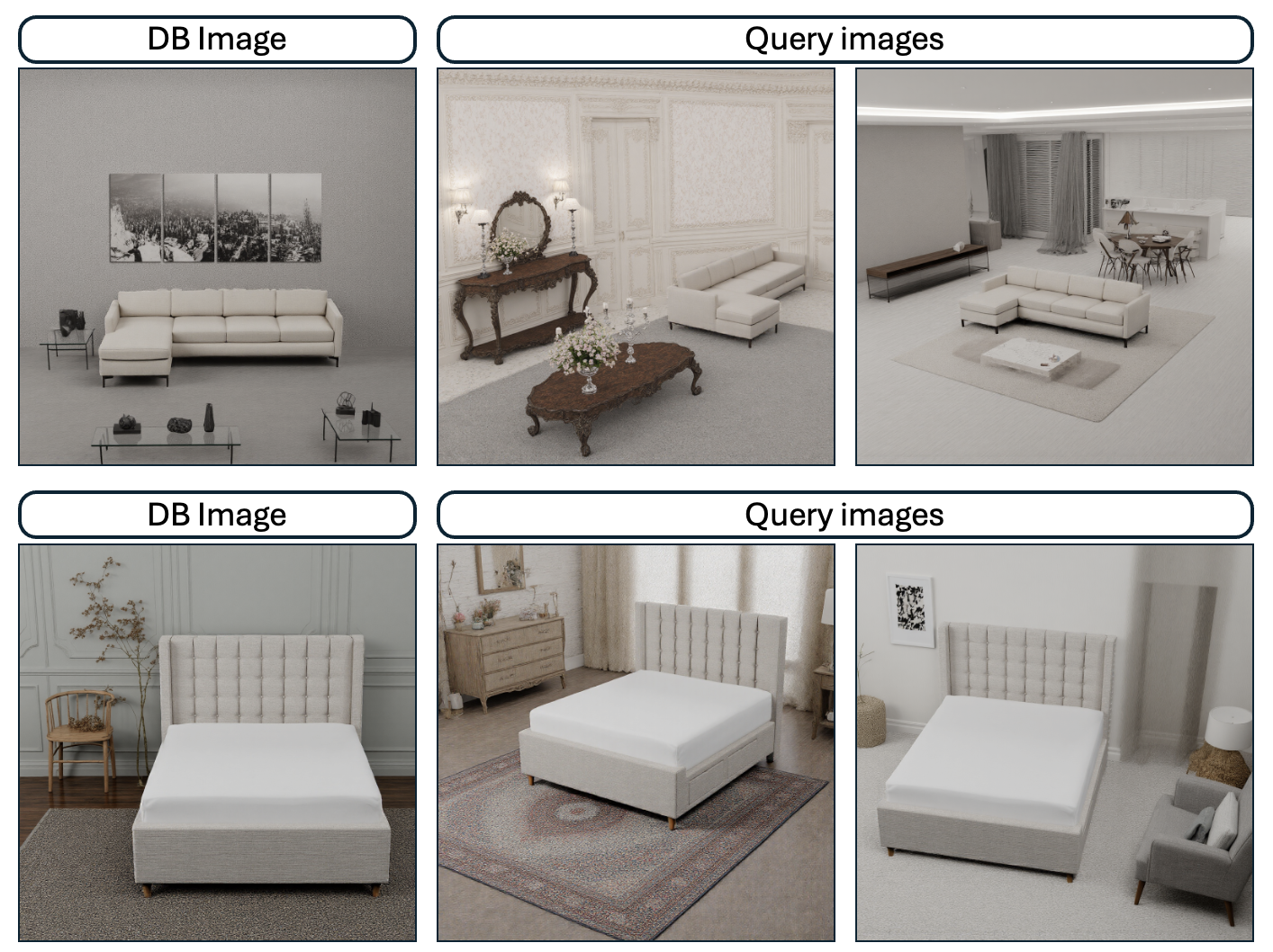}
    \caption{Generated samples are split into DB images (left) and query images (right).}
    \label{fig:generated-samples}
\end{figure*}

DB-query pairs that are synthesized from the same source item form ground-truth correspondences (see examples in Fig.~\ref{fig:generated-samples}).

\subsection{Dataset Statistics}
Table~\ref{tab:dataset-summary} reports per-category counts of unique items (\emph{samples}) and resulting images, partitioned into database (DB) and query sets produced by our data-generation pipeline.

\begin{table*}[h!]
\caption{Dataset summary by category. Each item is generated via our data-generation pipeline and split into database (DB) and query images.}
\label{tab:dataset-summary}
\centering
\small
\setlength{\tabcolsep}{7pt}
\resizebox{\textwidth}{!}{%
\begin{tabular}{lrrrr}
\toprule
Category & Samples & Images & DB images & Query images \\
\midrule
Bed      & 22  & 2939 & 1128 & 1811 \\
Cabinet  & 100 & 1007 &  201 &  806 \\
Chair    & 100 & 1005 &  201 &  804 \\
Desk     & 100 & 1001 &  200 &  801 \\
Dresser  & 62  &  624 &  272 &  352 \\
Planter  & 135 & 1349 &  667 &  682 \\
Shelf    & 100 & 1001 &  200 &  801 \\
Sofa     & 22  & 1652 &  630 & 1022 \\
Vase     & 50  &  501 &  250 &  251 \\
\midrule
\textbf{Total} & \textbf{691} & \textbf{11079} & \textbf{3749} & \textbf{7330} \\
\bottomrule
\end{tabular}}
\end{table*}

\subsection{Synthetic Data Samples}

\begin{figure}[h!]
    \centering
    \includegraphics[width=0.95\textwidth]{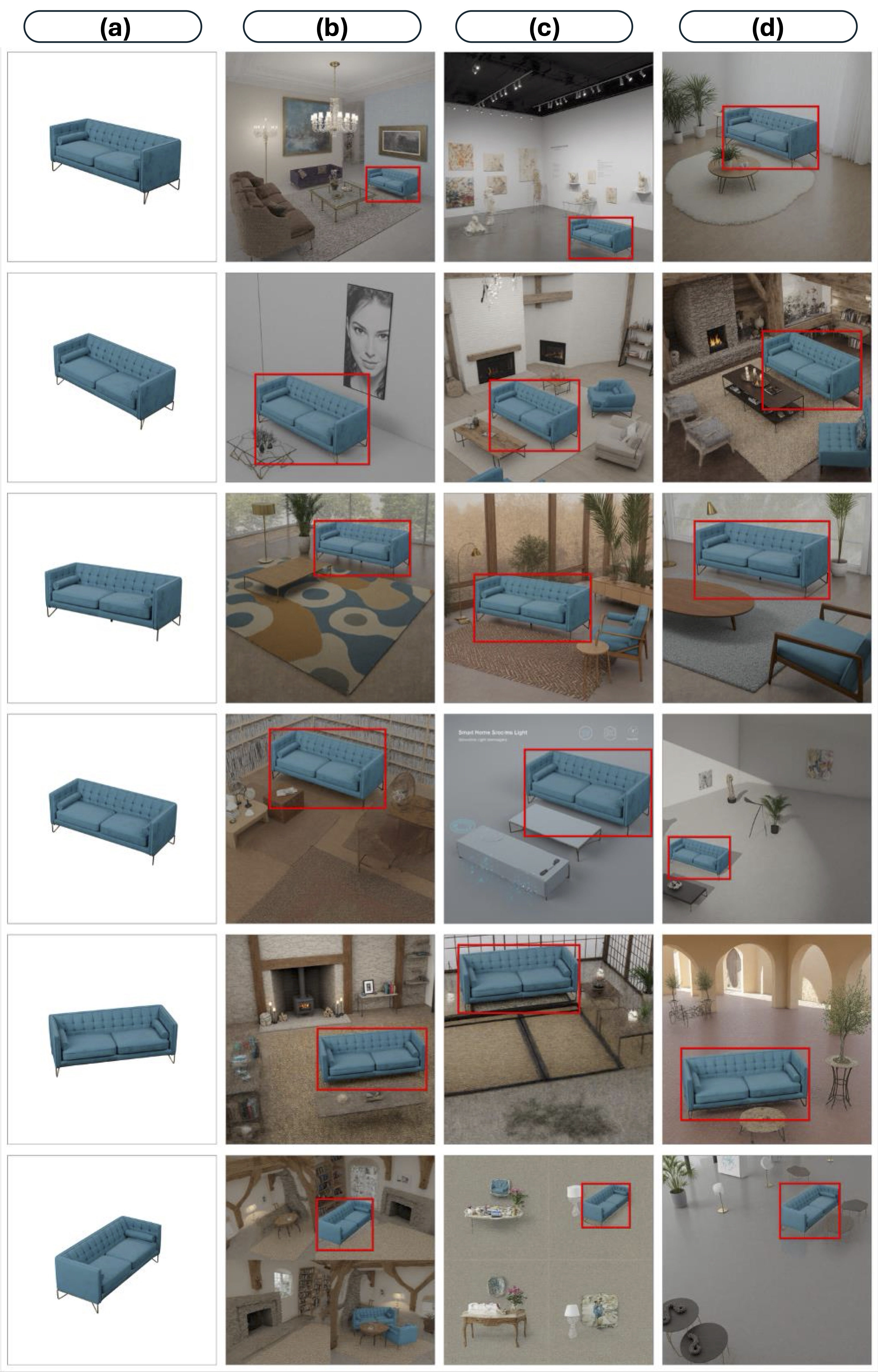}
    \caption{Generated data for a teal upholstered sofa with button-tufted back cushions and metal legs. Column (a) shows the original object in different rotations in isolation, while columns (b), (c) and (d) present the same object in the corresponding rotation (same row) scaled, shifted and integrated into various indoor scenes. The scenes include diverse residential and commercial environments such as modern living rooms, art galleries, minimalist spaces, rustic interiors with fire places, and contemporary office settings. Columns (b), (c), (d) also have the ground truth bounding box shown in red.} 
    \label{fig:data_generation_sofa}
\end{figure}

\begin{figure}[h!]
    \centering
    \includegraphics[width=0.95\textwidth]{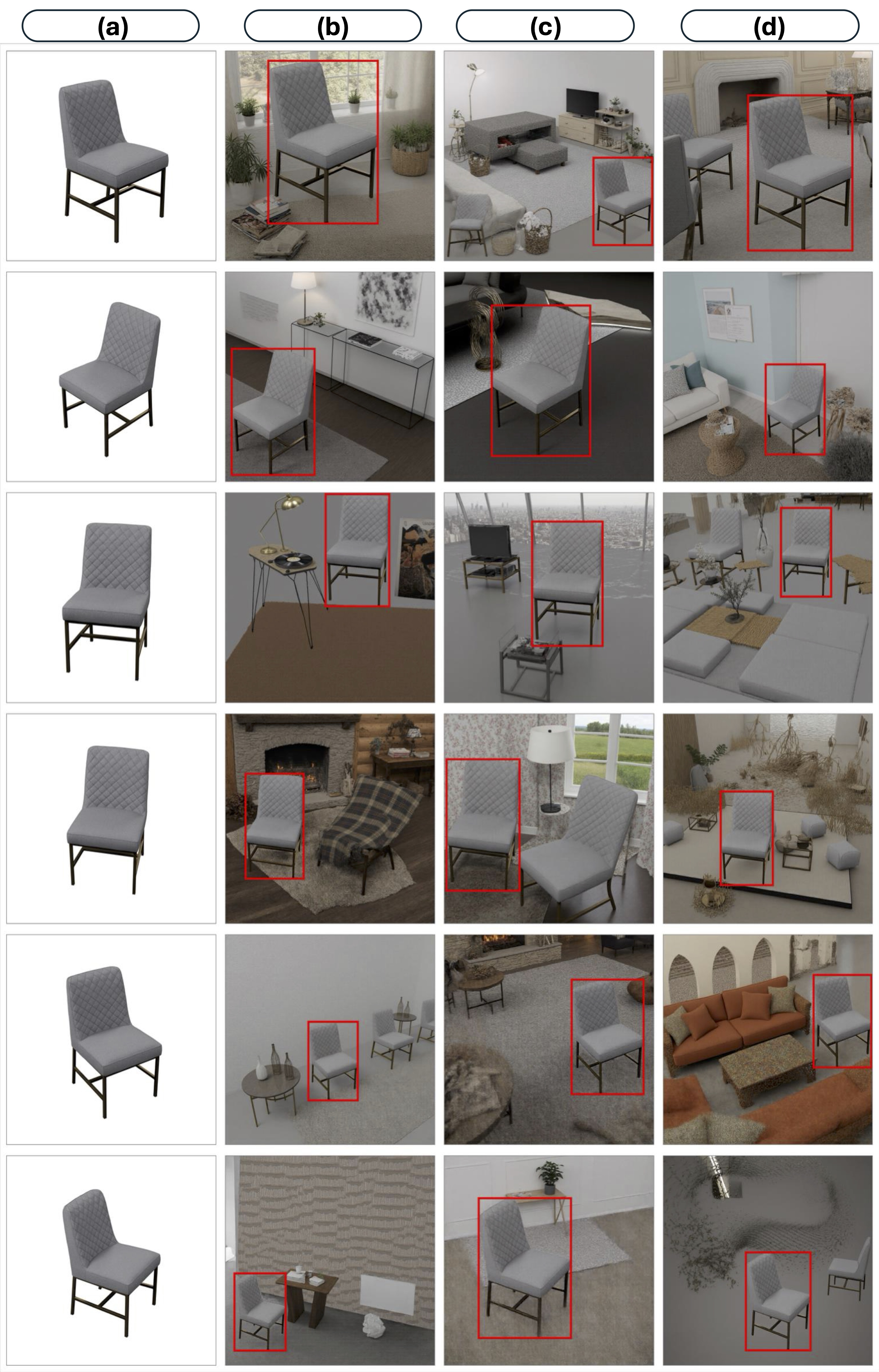}
    \caption{Generated data for a grey upholstered dining chair with quilted diamond-pattern backrest and black metal frame. Column (a) shows the original object in different rotations in isolation, while columns (b), (c) and (d) present the same object in the corresponding rotation (same row) scaled, shifted and integrated into various indoor scenes. The scenes include diverse residential and commercial environments such as modern living rooms, minimalist offices, rustic interiors with fireplaces, contemporary dining spaces, and gallery-like settings. Columns (b), (c), (d) also have the ground truth bounding box shown in red.} 
    \label{fig:data_generation_chair}
\end{figure}

\begin{figure}[h!]
    \centering
    \includegraphics[width=0.95\textwidth]{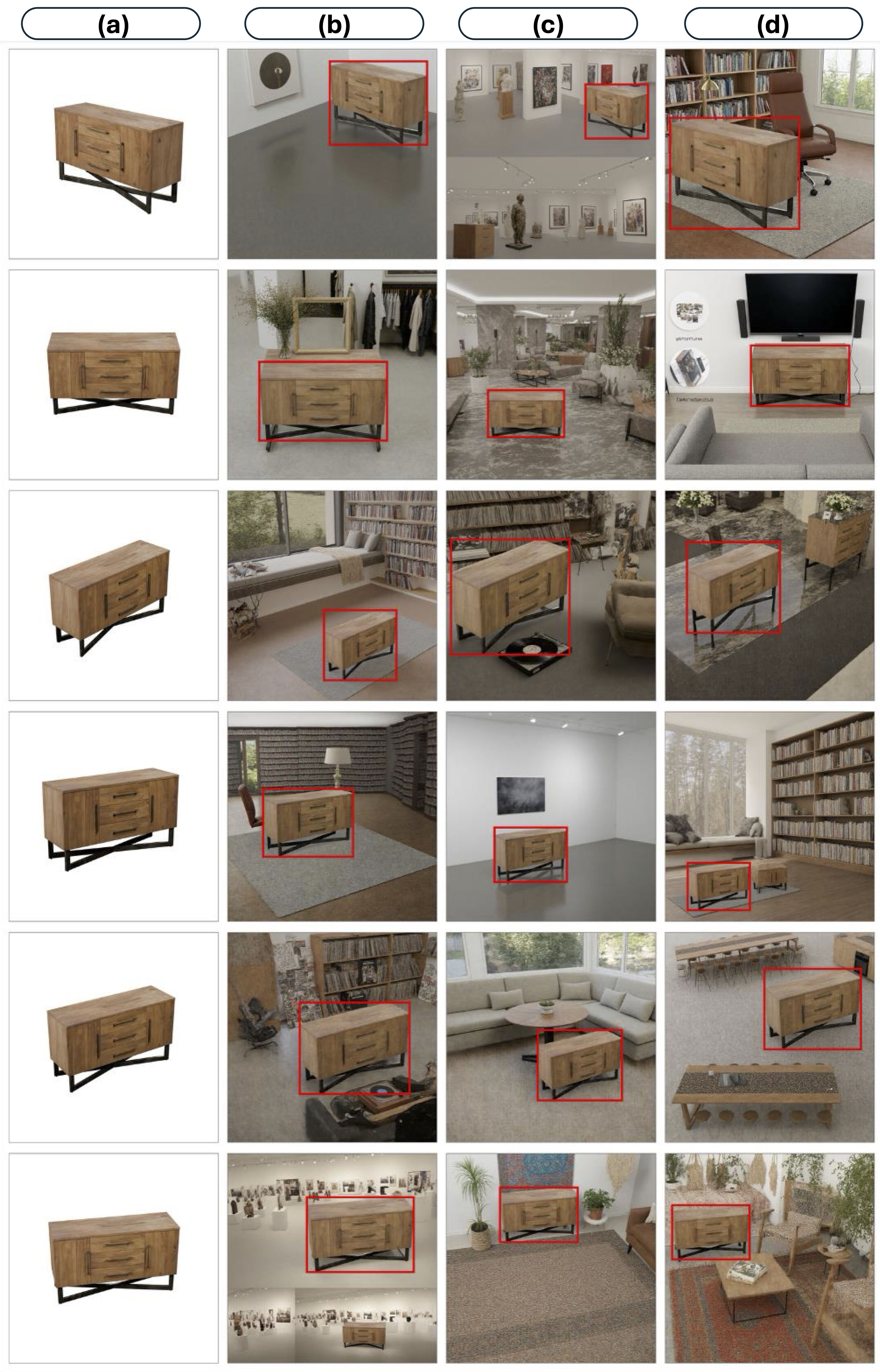}
    \caption{Generated data for a wooden sideboard with three drawers and black metal geometric frame base. Column (a) shows the original object in different rotations in isolation, while columns (b), (c) and (d) present the same object in the corresponding rotation (same row) scaled, shifted and integrated into various indoor scenes. The scenes include diverse residential and commercial environments such as modern living rooms, home offices with bookshelves, gallery spaces, contemporary bedrooms, minimalist interiors, and traditional spaces with brick walls. Columns (b), (c), (d) also have the ground truth bounding box shown in red.} 
    \label{fig:data_generation_cabinet}
\end{figure}

\begin{figure}[h!]
    \centering
    \includegraphics[width=0.95\textwidth]{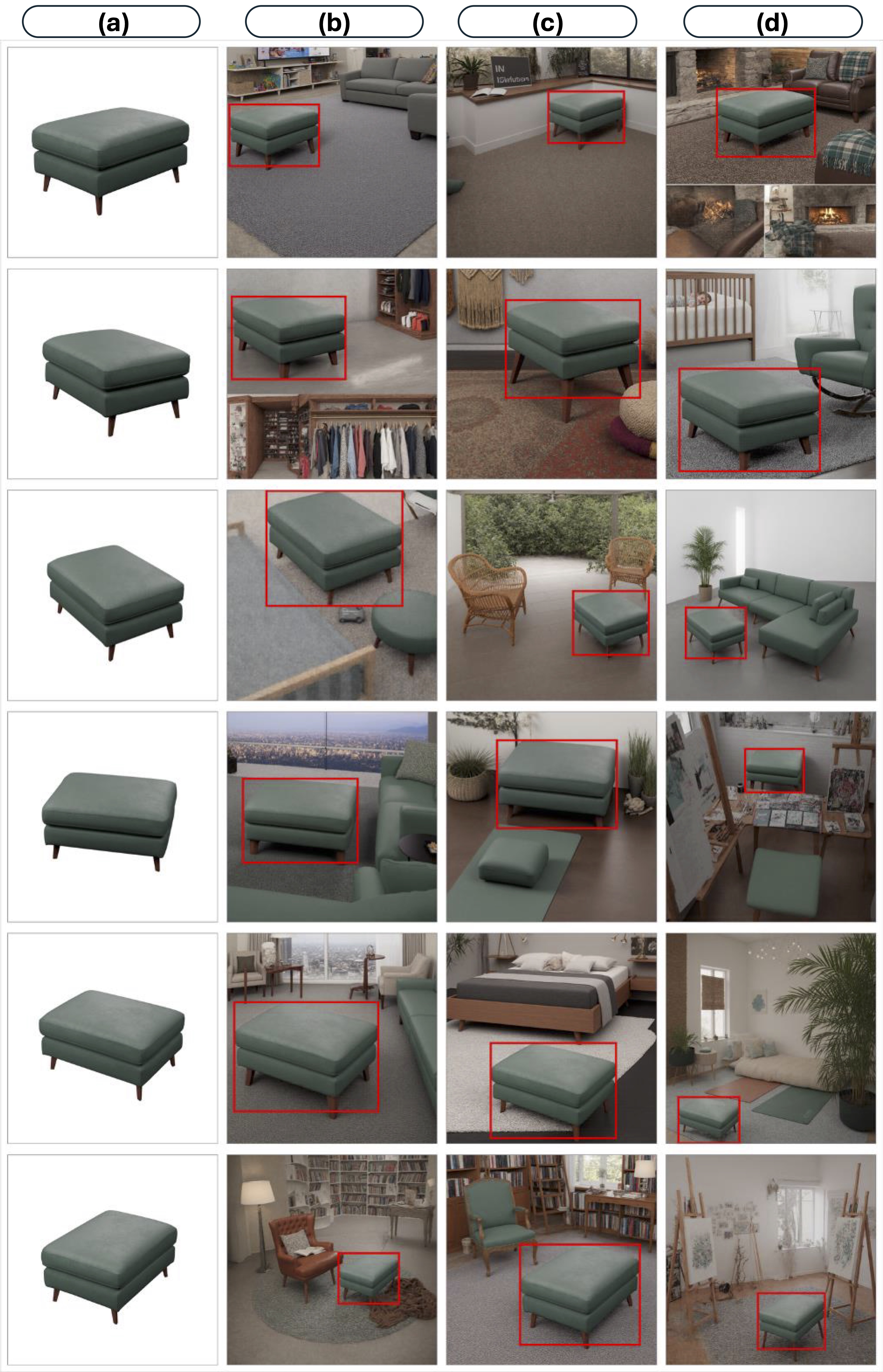}
    \caption{Generated data for a sage green upholstered ottoman with wooden legs. Column (a) shows the original object in different rotations in isolation, while columns (b), (c) and (d) present the same object in the corresponding rotation (same row) scaled, shifted and integrated into various indoor scenes. The scenes include diverse residential environments such as modern living rooms, cozy bedrooms, contemporary home offices, outdoor terraces, bohemian-style spaces with macramé decor, and traditional libraries with bookshelves. Columns (b), (c), (d) also have the ground truth bounding box shown in red.} 
    \label{fig:data_generation_ottoman}
\end{figure}

In this section, we show a few examples generated through our designed pipeline. Figures \ref{fig:data_generation_sofa}, \ref{fig:data_generation_chair}, \ref{fig:data_generation_cabinet} and \ref{fig:data_generation_ottoman} show data generated from different 3D objects in a multitude of everyday environments.

\clearpage
\section{Methodology}

\subsection{Model Groups}
\label{appendix:model-groups}

We evaluate two model families distinguished by output type and prompting interface: Task-specific vision models, and general-purpose VLMs. Table~\ref{tab:models-summary} provides a summary of model properties and prompting inference for the selected models.

\begin{table}[h]
\centering
\caption{Models evaluated in this study. (REC = referring expression comprehension).}
\label{tab:models-summary}
\small
\setlength{\tabcolsep}{6pt}
\resizebox{\textwidth}{!}{%
\begin{tabular}{l l l c c}
\toprule
 & \textbf{Model} & \textbf{Family / Style} & \textbf{Inference} & \textbf{Year} \\
\midrule
\multirow{7}{*}{\rotheadW[2.5cm]{\mbox{Task-specific}\\\mbox{vision models}}}
& D-FINE & Strong non-OVD detector & non-REC & 2024 \\
& OWL-ViT-base-32 & Promptable OVD & Word & 2022 \\
& OWLv2-base-16 & Promptable OVD (self-training) & Word & 2023 \\
& Florence-2-base & Foundation model used for detection & Sentence & 2024 \\
& Grounding DINO & Grounding-style pretraining & Word & 2023 \\
& LISA-7B & Reasoning Segmentation & Sentence & 2023 \\
& LISA-13B-Llama & Reasoning Segmentation & Sentence & 2023 \\
\midrule
\multirow{7}{*}{\rotheadW[2.5cm]{\mbox{General-purpose}\\VLMs}}
& LLaVA-1.5-7B & Open LVLM & Sentence & 2023 \\
& LLaVA-Next-7B & Open LVLM & Sentence & 2024 \\
& LLaVA-OneVision-7B-si & Open LVLM & Sentence & 2024 \\
& SmolVLM2-2.2B-Instruct & Open LVLM (efficient) & Sentence & 2024 \\
& InternVL3-8B & Open LVLM & Sentence & 2023 \\
& Gemini~2.5-Pro & API LVLM & Sentence & 2025 \\
& GPT-4o & API LVLM & Sentence & 2024 \\
\bottomrule
\end{tabular}}
\end{table}

\subsection{Evaluation Protocol and Metrics}
\label{appendix:eval-protocol}
We designed the following complementary tasks for benchmarking the performance of task-specific vision models and VLMs: 
\begin{itemize}
    \item \textbf{Localization (box-based):} Given an image and target name (e.g., \emph{sofa}), predict one box for the main product; task-specific models directly output class-conditioned bounding box, while VLMs are prompted in REC style to output a box (first valid output is used). Evaluation metrics include $mIoU$, $AP_{0.5}$, and $AP_{0.75}$, $AP_{0.5{:}0.95}$. 
    \item \textbf{Localization (coarse grid):} We overlay a $2{\times}2$ or $3{\times}3$ grid and define the GT cell by majority overlap with the GT box. VLMs return a cell index/tag; For task=specific vision models, top-1 predicted boxes are converted by assigning the cell with maximum overlap. We use multiclass metrics: accuracy, macro-$\mathrm{F}_{1}$, and multiclass MCC. 
    \item \textbf{Spatial reasoning -- coarse depth ordering:} VLMs are prompted to answer with \{\texttt{front}, \texttt{back}\} under a constrained prompt whether the focal object is in front or back of the scene; we report accuracy, precision, recall, and $\mathrm{F}_{1}$.
    \item \textbf{Patch-based retrieval (downstream):} We crop the predicted box (clipped to bounds), embed the crop with VL-CLIP~\cite{vlclip2025}, and query an HNSW index~\cite{hnsw2018} built over DB embeddings to return top-$k$ candidates. Models are compared using Precision@$k$ and Hit@$k$ for $k\in\{1,2,3\}$, based on the DB–query GT correspondences obtained through the data synthesis pipeline. 
\end{itemize}

\section{Results}
\label{appendix:results}

\subsection{Localization}

Table~\ref{tab:loc-combined} provides the box-based localization performance for both task-specific vision models and general-purpose VLMs. Among task-specific models, GroundingDINO leads, followed by Florence-2 and OWLv2. Notably, while D-FINE attains strong mIoU, its AP is the lowest in this group, indicating a calibration gap between box quality and confidence scores under our setup.

Among general-purpose VLMs, InternVL is the strongest, followed by LLaVA; while other VLMs fall off rapidly, especially at stricter IoU thresholds where AP nearly collapses.

\begin{table*}[h!]
\caption{Localization results (box-based) for task-specific vision models and general-purpose VLMs.}
\label{tab:loc-combined}
\centering
\small
\setlength{\tabcolsep}{7pt}
\resizebox{\textwidth}{!}{%
\begin{tabular}{l l c c c c}
\toprule
 & Model & mIoU~$\uparrow$ & $AP_{0.5}$~$\uparrow$ & $AP_{0.75}$~$\uparrow$ & $AP_{0.5:0.95}$~$\uparrow$ \\
\midrule
\multirow{7}{*}{\rotheadW[2.5cm]{\mbox{Task-specific}\\\mbox{vision models}}}
& GroundingDINO-1.5   & \textbf{0.773} & \textbf{0.821} & \textbf{0.711} & \textbf{0.695} \\
& Florence2-base      & 0.767          & \underline{0.755} & 0.617          & \underline{0.602} \\
& OWLv2-base-16       & 0.735          & 0.743            & \underline{0.631} & 0.592 \\
& LISA-13B-Llama      & 0.711          & 0.658            & 0.513          & 0.505 \\
& LISA-7B             & 0.649          & 0.535            & 0.427          & 0.409 \\
& OWL-ViT-base-32     & 0.706          & 0.436            & 0.374          & 0.338 \\
& D-FINE              & \underline{0.768} & 0.340          & 0.320          & 0.309 \\
\midrule
\multirow{7}{*}{\shortstack{\rotheadW[2.5cm]{\mbox{General-purpose}\\\mbox{VLMs}}}}
& InternVL3-8B                & \textbf{0.550} & \textbf{0.454} & \underline{0.079} & \underline{0.160} \\
& LLaVA-Next-7B               & \textbf{0.550} & 0.342          & \textbf{0.158}    & \textbf{0.173} \\
& LLaVA-1.5-7B                & 0.487          & \underline{0.389} & 0.022          & 0.111 \\
& LLaVA-OneVision-7B-si       & 0.388          & 0.148          & 0.001          & 0.029 \\
& Gemini~2.5-Pro              & 0.210          & 0.010          & 0.000          & 0.002 \\
& GPT-4o                      & 0.181          & 0.003          & 0.000          & 0.000 \\
& SmolVLM2-2.2B-Instruct      & 0.136          & 0.005          & 0.000          & 0.000 \\
\bottomrule
\end{tabular}}
\end{table*}

Based on results in Table~\ref{tab:loc-combined}, task-specific models substantially outperform general-purpose VLMs on both mIoU and AP across different thresholds. Practically, this suggests a simple hybrid strategy for applications that need both reasoning and high-accuracy boxes: first leverage a reasoning-capable VLM to resolve the referent via language (e.g., infer the product/object name or category), then feeding the VLM output  to a specialized open-vocabulary detector (OVD) for precise localization. This ``reason-then-localize'' pipeline preserves the VLM's scene understanding while delegating box regression and confidence calibration to models optimized for detection.

\begin{figure*}[h!]
    \centering
    \includegraphics[width=1\linewidth]{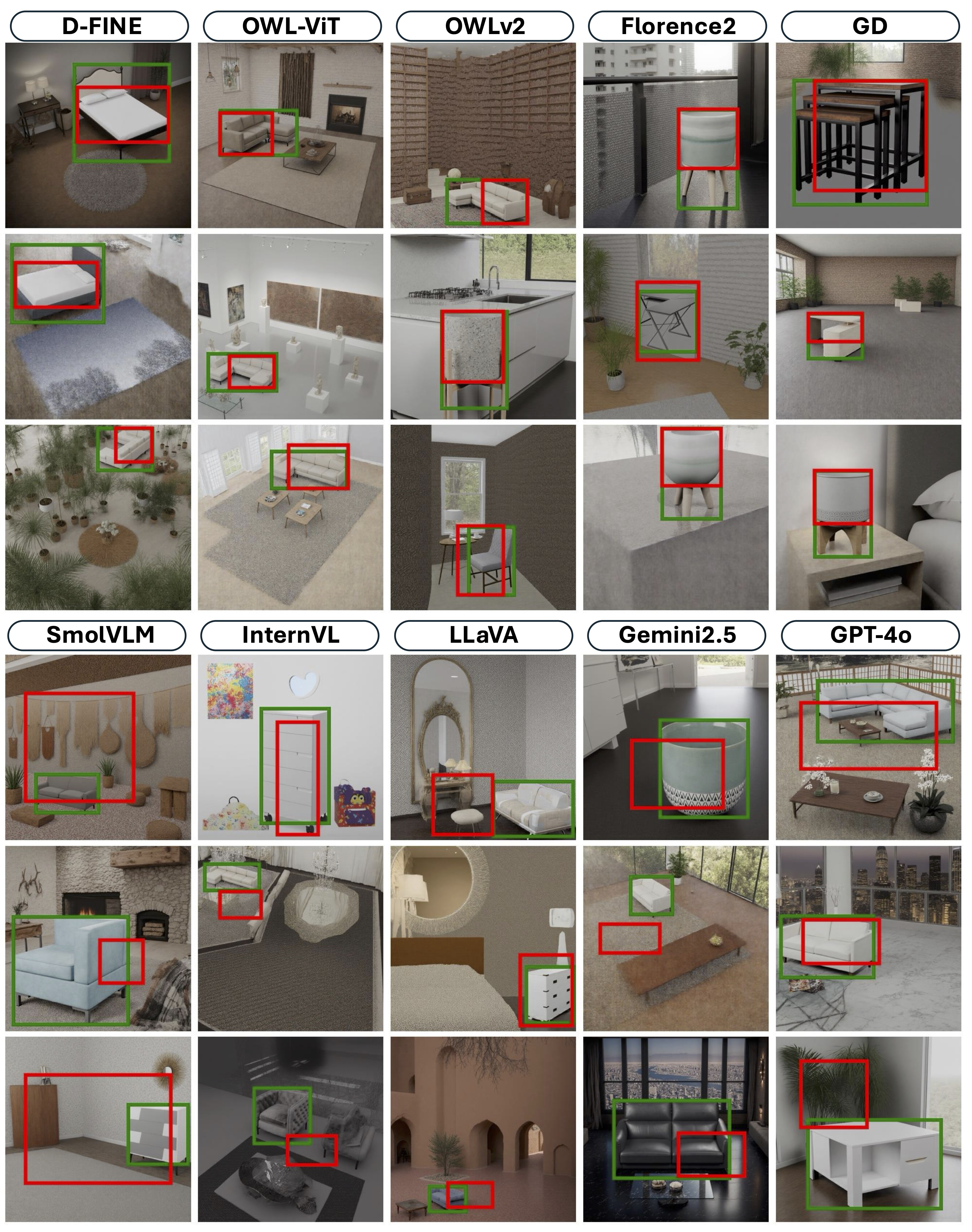}
    \caption{\textbf{Spatial localization failure cases across task-specific vision models (top) and general-purpose VLMs (bottom).}
    Green boxes denote ground truth; red boxes are predictions.} 
    \label{fig:spatial-failure-cases}
\end{figure*}

\subsection{Spatial Localization Analysis}
\label{appendix:spatial-analysis}

Figure~\ref{fig:grid-failures} illustrates typical errors from five VLMs when predicting coarse grid locations of the focal product. 
We show two examples per model for both a \(2\times2\) grid and a \(3\times3\) grid. 
The ground-truth and model’s prediction cells are marked with {green} and red symbols.
Errors are often adjacent-cell mistakes near cell boundaries.
Moving from $2\times 2$ to $3\times 3$ grid increases difficulty and error frequency due to finer spatial quantization and ambiguity in cluttered scenes.

\subsection{Retrieval Analysis}
\label{appendix:ann-retrieval-results}

\begin{table*}[h!]
\caption{ANN retrieval performance over VL-CLIP embeddings using predicted-box crops. The first row shows the full-image (no crop) as a baseline.}
\label{tab:retrieval-all}
\centering
\small
\resizebox{\textwidth}{!}{
\begin{tabular}{l l c c c c c}
\toprule
 & & \multicolumn{3}{c}{Precision@$k$~$\uparrow$} & \multicolumn{2}{c}{Hit@$k$~$\uparrow$} \\
\cmidrule(lr){3-5} \cmidrule(lr){6-7}
 & Model & @1 & @5 & @10 & @5 & @10 \\
\midrule
 & Full image (no crop) & 0.400 & 0.285 & 0.236 & 0.576 & 0.659 \\
\midrule
\multirow{7}{*}{\rotheadW[2.5cm]{\mbox{Task-specific}\\\mbox{vision models}}}
& GroundingDINO-1.5           & \textbf{0.554} & \textbf{0.426} & \textbf{0.270} & \textbf{0.736} & \textbf{0.790} \\
& OWL-ViT-base-32-base-32     & 0.551 & \textbf{0.426} & \underline{0.266} & 0.721 & 0.768 \\
& Florence-2-base             & \underline{0.539} & 0.419 & 0.264 & \underline{0.728} & \underline{0.788} \\
& OWLv2-base-16-base-16       & 0.517 & 0.404 & 0.256 & 0.699 & 0.759 \\
& LISA-13B-Llama              & 0.499 & 0.388 & 0.248 & 0.686 & 0.737 \\
& LISA-7B                     & 0.467 & 0.363 & 0.231 & 0.643 & 0.693 \\
& D-FINE                      & 0.447 & 0.310 & 0.251 & 0.581 & 0.633 \\
\midrule
\multirow{7}{*}{\rotheadW[2.5cm]{\mbox{General-purpose}\\VLMs}}
& LLaVA-Next-7B               & \textbf{0.498} & \textbf{0.391} & \textbf{0.244} & \textbf{0.671} & \textbf{0.725} \\
& LLaVA-1.5-7B                & \underline{0.448} & 0.344 & 0.225 & \underline{0.634} & \underline{0.714} \\
& InternVL3-8B                & 0.445 & \underline{0.346} & \underline{0.228} & 0.625 & 0.700 \\
& LLaVA-OneVision-7B-si       & 0.345 & 0.279 & 0.185 & 0.526 & 0.605 \\
& SmolVLM2-2.2B-Instruct      & 0.277 & 0.209 & 0.147 & 0.421 & 0.491 \\
& Gemini~2.5-Pro              & 0.173 & 0.135 & 0.096 & 0.254 & 0.308 \\
& GPT-4o                      & 0.129 & 0.112 & 0.081 & 0.233 & 0.281 \\
\bottomrule
\end{tabular}}
\end{table*}

We evaluate ANN retrieval after cropping each query image to the predicted bounding box from each model and embedding the crop with VL-CLIP. We report retrieval performance using Precision@$k$ and Hit@$k$ for $k\in{1,5,10}$ compared against a full-image baseline (Table~\ref{tab:retrieval-all}).

GroundingDINO yields the strongest overall retrieval among the task-specific models. All models in this group improve over the full-image baseline, although D-FINE slightly underperforms the baseline on Hit rate. 

LLaVA-Next is the top performer across all metrics among general-purpose VLMs, while Gemini and GPT-4o show the lowest performance, falling below the no-crop baseline.

\paragraph{Retrieval failure cases}

\begin{figure*}[h]
    \centering
    \includegraphics[width=\linewidth]{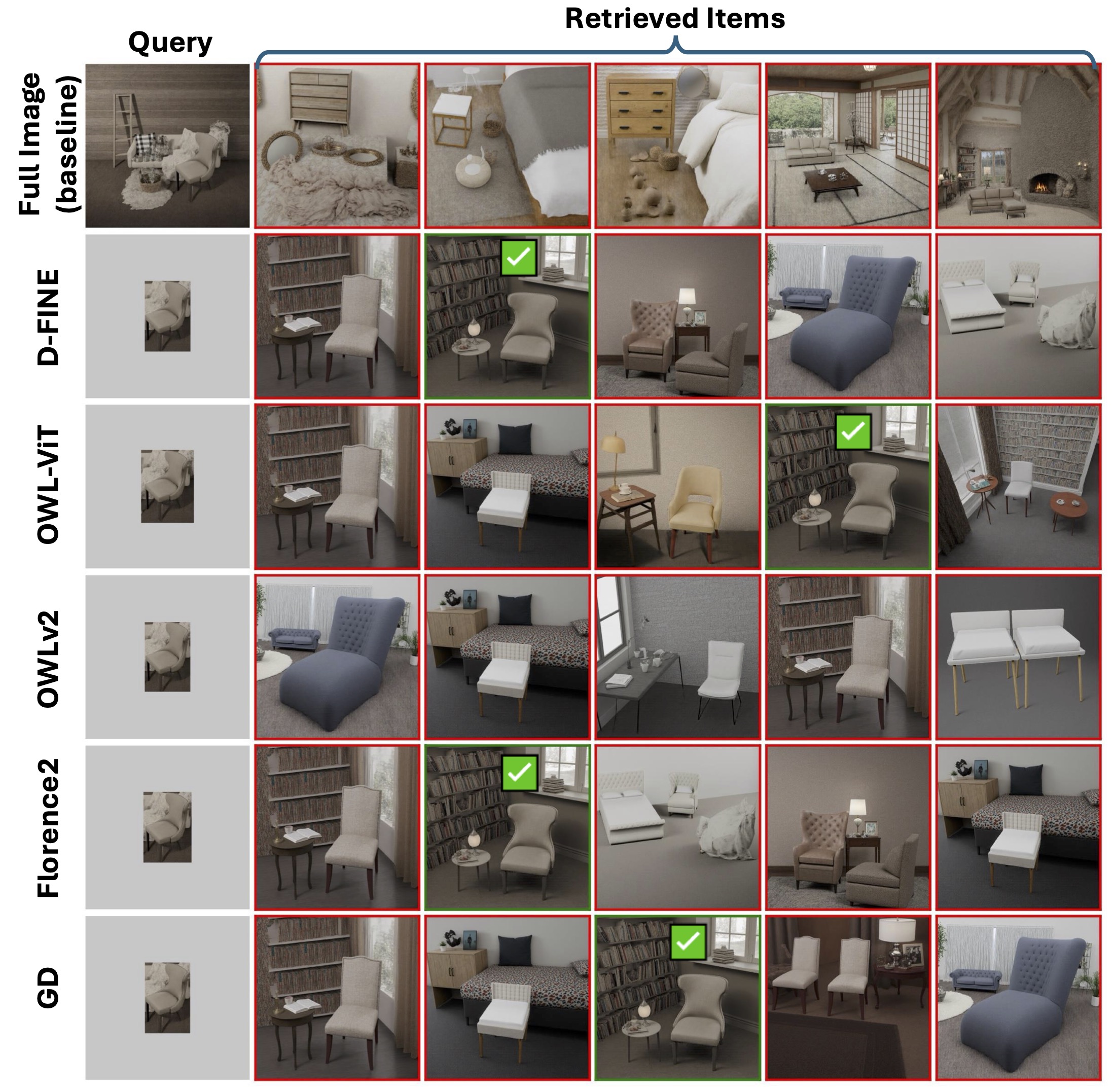}
    \caption{\textbf{Retrieval results with crops made from predicted boxes of task-specific vision models (top-5 shown).}
    Row~1: full-image baseline (no crop). Rows~2--6: crops from D-FINE, OWL-ViT, OWLv2, Florence-2, and GroundingDINO, respectively. 
    The \emph{same query image} is used across all rows; only the crop differs by model. 
    In this example, the baseline and OWLv2 produce \emph{no} correct matches in the top-5; \emph{D-FINE} and Florence-2 retrieve a correct match at rank~2; OWL-ViT and GroundingDINO retrieve a correct match at rank~3. 
    This highlights how modest crop shifts (e.g., tighter/looser boxes or slight offsets) can substantially alter retrieval outcomes.}
    \label{fig:retrieval-failure-cases-p1}
\end{figure*}

\begin{figure*}[h]
    \centering
    \includegraphics[width=\linewidth]{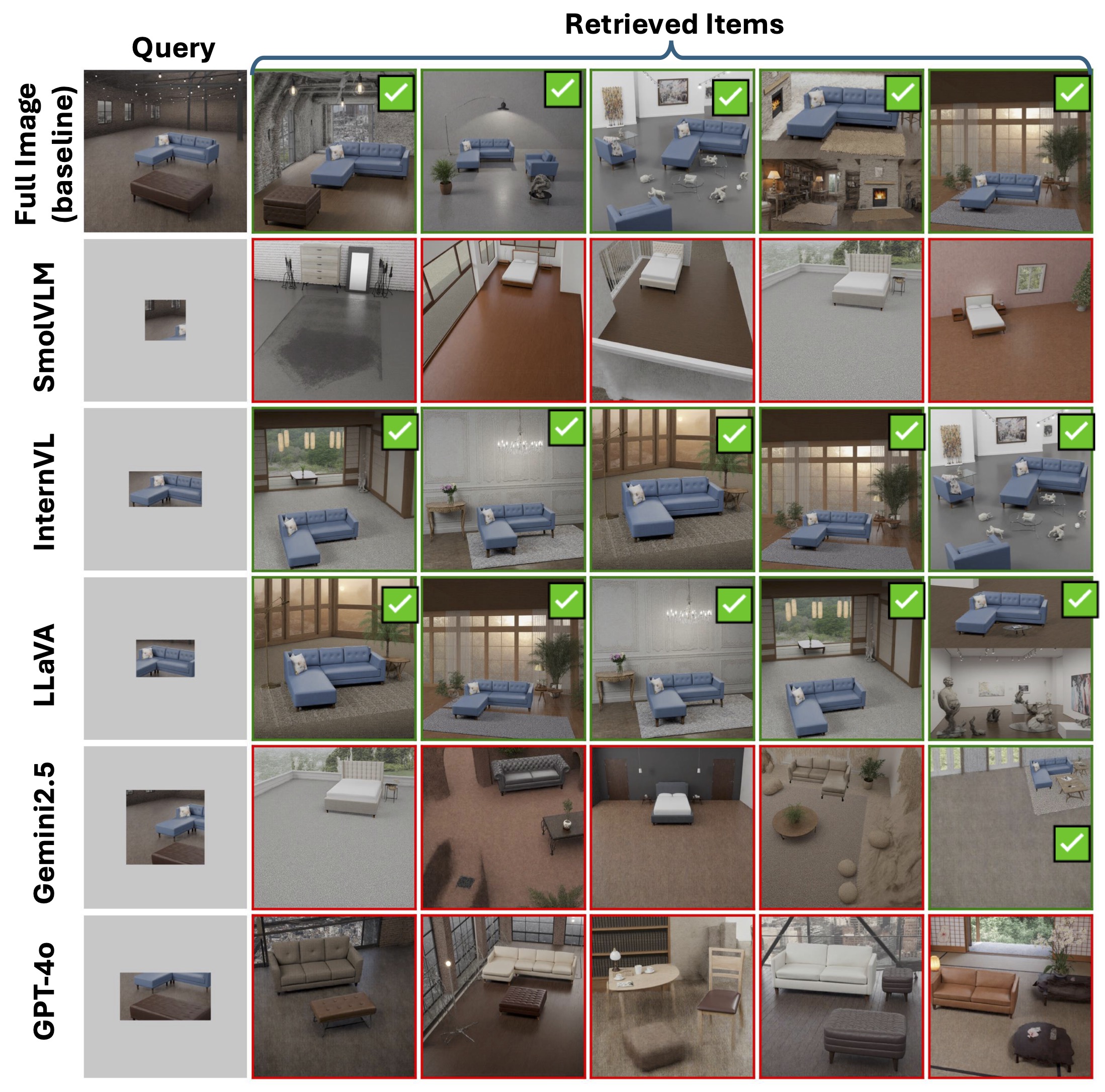}
    \caption{\textbf{Retrieval results using crops made from predicted boxes of general-purpose VLMs (top-5 shown).}
    Row~1: full-image baseline. Rows~2--6: crops from SmolVLM, InternVL, LLaVA, Gemini~2.5, and GPT-4o, respectively. 
    The \emph{same query image} is used across all rows. 
    Here, the full-image baseline, InternVL, and LLaVA achieve \emph{all} top-5 correct; SmolVLM and GPT-4o have \emph{no} correct matches; Gemini~2.5 produces a single correct match at rank~5.
    }
    \label{fig:retrieval-failure-cases-p2}
\end{figure*}

Figures~\ref{fig:retrieval-failure-cases-p1} and~\ref{fig:retrieval-failure-cases-p2} visualize top-5 ANN retrieval results using a single query image across models. 
Row~1 in each figure shows the full-image baseline (i.e., retrieval on full image without cropping). Subsequent rows use query crops from the predicted bounding boxes of each model, embedded with VL-CLIP for retrieval. 
Columns display retrieved candidates left-to-right (rank~1 to~5); correct matches are outlined with \textcolor{green!70!black}{green} and incorrect ones in \textcolor{red}{red} borders. 
Small differences in the predicted crop can induce significant changes in downstream retrieval. Models that better preserve salient, object-specific features in their boxes tend to yield better matching retrievals.

Results reinforce that crops which preserve the correct visual features (pose, discriminative parts) enable consistent instance-level retrieval, whereas suboptimal boxes cause retrieval to drift toward contextually similar but incorrect items.

\clearpage

\clearpage
\section{VLM Prompts}
\label{appendix:llm-prompts}

\subsection{Object position on a $2\times 2$ grid}
\label{appendix:4-grid-prompt}
The prompt used in the $2\times 2$ grid-cell localization experiment; the VLM selects the dominant quadrant (A–D) containing the target object.

\begin{tcolorbox}[promptbox]
\begin{lstlisting}[
  backgroundcolor=\color{black!0},
  basicstyle=\small\ttfamily\color{black},
]
Divide this image into 4 regions. Where is the {object} most prominently located in this image? If the object spans multiple regions, choose the region where the majority or most prominent part of the object is located.

Options:
A) top-left
B) top-right
C) bottom-left
D) bottom-right

Chose from one of the above options. 
\end{lstlisting}
\end{tcolorbox}

\subsection{Object position on a $3\times 3$ grid}
\label{appendix:9-grid-prompt}

The prompt used for the $3\times 3$ grid-cell localization experiment; the VLM selects the dominant cell (A–I) where the target object lies.

\begin{tcolorbox}[promptbox]
\begin{lstlisting}[
  backgroundcolor=\color{black!0},
  basicstyle=\small\ttfamily\color{black},
]
Divide this image into 9 regions. Where is the {object} most prominently located in this image? If the object spans multiple regions, choose the region where the majority or most prominent part of the object is located.

Options:
A) top-left
B) top-center
C) top-right
D) middle-left
E) middle-center
F) middle-right
G) bottom-left
H) bottom-center
I) bottom-right

Chose from one of the above options.
\end{lstlisting}
\end{tcolorbox}

\subsection{Object Depth Estimation}
\label{appendix:depth-prompt}
The prompt used in the front/back depth-order classification experiment; the VLM indicates whether the target object is in the foreground or background.

\begin{tcolorbox}[promptbox]
\begin{lstlisting}[
  backgroundcolor=\color{black!0},
  basicstyle=\small\ttfamily\color{black},
]
Look at the {object} in this image. Is it positioned in the foreground (front) or background (back) of the scene?
Consider the relative depth and layering of objects in the image.

A) Front (foreground)
B) Back (background)

Chose from one of the above options.
\end{lstlisting}
\end{tcolorbox}

\subsection{Object Bounding Box}
\label{appendix:bbox-prompt}
\begin{tcolorbox}[promptbox]
\begin{lstlisting}[
  backgroundcolor=\color{black!0},
  basicstyle=\small\ttfamily\color{black},
]
Look at the {object} in this image. Please identify the bounding box coordinates that tightly enclose this object.

Provide the coordinates as absolute pixel values based on the image dimensions:
- Image width: {width} pixels
- Image height: {height} pixels
- (0,0) is the top-left corner of the image

Format: (x1, y1, x2, y2) where:
- x1, y1 = top-left corner of the bounding box in pixels
- x2, y2 = bottom-right corner of the bounding box in pixels

Response format: Provide only the coordinates in the specified format.
\end{lstlisting}
\end{tcolorbox}

\section{Related Works}

\subsection{Object Detection: From Closed-set to Open-vocabulary}
Early object detectors were based on a fixed category list, often referred to as the closed-set detectors, where a model is trained and evaluated on the same finite taxonomy. This limitted inference applications: adding a new class required retraining.\cite{rcnn2014,fastrcnn2015,fasterrcnn2016,yolov12016,yolov8_ultralytics,detr2020,dino2022}  To relax the fixed-label constraint, open-vocabulary detection (OVD) brings text supervision into the loop. Early OVD lines either distilled knowledge from vision–language models into detectors or learned region–text alignment at scale.\cite{vild2021,regionclip2022,glip2022,glipv22022}  

Grounding DINO~\cite{gd2024} adopts a dual-encoder, single-decoder architecture with an image encoder, a language encoder, and a cross-modality decoder for box refinement. Its 1.5 variant~\cite{gd1pt52024} employs a larger ViT-L backbone~\cite{vit2020} and is pre-trained on over 20M grounded image–text pairs, reporting 54.3 AP on COCO and 55.7 AP on the LVIS \emph{minival} zero-shot transfer benchmark.
OWL-ViT~\cite{owlvit2022} pre-trains vision and text encoders with image–text contrastive learning (as in CLIP~\cite{clip2021} and ALIGN~\cite{align2021}) and adapts them for open-vocabulary detection with text prompts. OWL-ST and OWL-v2~\cite{owlv22023} further scale this approach via self-training, using an existing detector to generate pseudo-boxes on web-scale image–text pairs, yielding substantial gains on rare LVIS categories.

Florence~\cite{florence2021} pre-trains language and image encoders with contrastive objectives and adapts to detection by attaching a Dynamic Head adapter~\cite{dynamichead2021}. Its successor, Florence-2~\cite{florence22024}, comprises an image encoder and a multimodal encoder–decoder trained under a unified multi-task paradigm. While not language-grounded, D-FINE~\cite{dfine2024} revisits bounding-box regression with fine-grained distribution refinement and global localization self-distillation, offering competitive AP at high FPS.\cite{dfine2024} Complementary to box-based OVD, LISA~\cite{lisa2024} performs reasoning segmentation, predicting language-conditioned masks with an LLM-guided planner.\cite{lisa2024} Broader efforts further expand OVD with larger corpora, using semi/self-training, and efficiency-oriented designs.\cite{detclip2022,detclipv22023,detclipv32024,detic2022,fvlm2022,yoloworld2024}

\subsection{Vision–Language Foundation Models for Object Localization and Spatial Reasoning}
Modern vision–language models (VLMs) pair a visual encoder with a language model and learn from large image–text corpora. Common design choices include lightweight connectors (projection or cross-attention), instruction tuning for task following, and support for multi-image or video inputs. This line of work established a general recipe for multimodal reasoning and flexible prompting.\cite{flamingo2022,openflamingo2023,blip22023,instructblip2023,lxmert2019,idefics22024,idefics32024}

LLaVA (Large Language and Vision Assistant)~\cite{llava2023,llava1pt52024} and its successors illustrate the rapid progress in open multimodal assistants. LLaVA connects a CLIP vision encoder to a Vicuna-based LLM via a simple projection and is trained end-to-end through a two-stage instruction-tuning pipeline~\cite{llava2023}, achieving state-of-the-art accuracy on ScienceQA and demonstrating strong visual dialogue abilities. 
SmolVLM~\cite{smolvlm2025} is a compact model that can run on-device capable of performing tasks such as visual question answering, captioning, and visual storytelling.

Recent VLM research also explores localization and dense description. InternVL~\cite{internvl2024}, Gemini~\cite{gemini2pt5pro2025}, GPT-4o~\cite{gpt4o2024}, and Gemma~3~\cite{gemma32025} all support localization via bounding boxes or segmentation. As VLMs continue to grow in scale and multimodality, they increasingly unify tasks such as visual question answering, open-ended captioning, object localization, and more, moving the field beyond simple image captioning toward general-purpose vision--language understanding.

\section{Discussion and Future Works}
We introduced a unified benchmark and evaluation protocol for product-centric retrieval that bridges detection and instance-level matching. Using a synthetic data pipeline, each product yields database images with more frontal views and query images with more angled views, enabling controlled tests of view/pose robustness. We index database embeddings with VL-CLIP and evaluate localization performance of task-specific vision models (e.g., OWL-ViT, GroundingDINO, D-FINE) alongside general purpose VLM (e.g., LLaVA, SmolVLM, InternVL). Localization quality (mIoU, AP) and retrieval quality (Precision@k, Hit@k for $k\in${1,5,10}) are measured under a common setup.
Across experiments, precise crops are a primary driver of retrieval success: using whole-image queries amplifies background bias, while missed/imprecise boxes and severe view changes are the dominant failure modes. 

Overall, our analysis demonstrates that task-specific vision models consistently outperform general-purpose VLMs accross all experiments. 
Future directions of this work includes (i) end-to-end training that jointly optimizes localization and retrieval embeddings, (ii) stronger view- and pose-invariant representations (e.g., 3D/geometry cues or multi-view augmentation), (iii) spatial reasoning over multi-object scenes (compositional relations and complements), and (iv) scaling to richer real-world catalogs with harder negatives and human-in-the-loop evaluation.

\end{document}